\documentclass[conference]{ieeeconf}
\IEEEoverridecommandlockouts
\usepackage{cite}
\usepackage{amsmath,amssymb,amsfonts}
\usepackage{algpseudocode}
\usepackage{graphicx}
\usepackage{textcomp}
\usepackage{algorithm}
\usepackage{float}
\usepackage{xcolor}
\usepackage{url}
\usepackage{booktabs}
\usepackage{multirow}
\usepackage{flushend}
\def\BibTeX{{\rm B\kern-.05em{\sc i\kern-.025em b}\kern-.08em
    T\kern-.1667em\lower.7ex\hbox{E}\kern-.125emX}}

\begin{document}

\title{Experimental Characterization of Fingertip Trajectory Following 
for a 3-DoF Series-Parallel Hybrid Robotic Finger}

\author{ Nicholas Baiata and Nilanjan Chakraborty
\thanks{This work was partially supported by the US Department of Defense through ALSRP 
under award No.\ HT94252410098, a SBU OVPR award, and a SBU LINCATS award.
Nicholas Baiata was supported by a Department of Education GAANN fellowship.}
\thanks{
The authors are with the Department of Mechanical Engineering, Stony Brook University, NY, USA. 
Emails: \{nicholas.baiata, nilanjan.chakraborty\}@stonybrook.edu.}}

\maketitle

\begin{abstract}
Task-space control of robotic fingers is a critical enabler of dexterous manipulation, as manipulation objectives are most naturally specified in terms of fingertip motions and applied forces rather than individual joint angles. While task-space planning and control have been extensively studied for larger, arm-scale manipulators, demonstrations of precise task-space trajectory tracking in compact, multi-DoF robotic fingers remain scarce. In this paper, we present the physical prototyping and experimental characterization of a three-degree-of-freedom, linkage-driven, series–parallel robotic finger with analytic forward kinematics and a closed-form Jacobian. A resolved motion rate control (RMRC) scheme is implemented to achieve closed-loop task-space trajectory tracking. We experimentally evaluate the fingertip tracking performance across a variety of trajectories, including straight lines, circles, and more complex curves, and report millimeter-level accuracy. To the best of our knowledge, this work provides one of the first systematic experimental demonstrations of precise task-space trajectory tracking in a linkage-driven robotic finger, thereby establishing a benchmark for future designs aimed at dexterous in-hand manipulation.
\end{abstract}


\section{INTRODUCTION}
Task-space control is a cornerstone of modern robotics because it allows specifying and executing motions directly in terms of end-effector positions and orientations, which are quantities most relevant to manipulation tasks. In dexterous manipulation, we are rarely interested in individual joint angles; rather, we care about applying forces, displacements, and velocities at specific points on the fingertips or the grasped object. Task-space control provides a natural interface for such objectives, coordinating multiple joints to achieve precise fingertip motions and enabling higher-level planners to operate in the space where tasks are specified~\cite{Nakanishi2008OperationalSpace}. Despite its importance, systematic demonstrations of task-space trajectory tracking remain rare in the context of robotic hands. Therefore, in this paper, \emph{we are interested in the experimental characterization of task-space trajectory tracking capability of a $3$-degree-of-freedom (DoF) robotic finger with both extension/flexion and abduction/adduction capability.}


Despite the wide array of actuation types (direct, tendon, linkage, pneumatic, shape memory polymers) and countless designs within each actuation type (see~\cite{kashef2020robotic, piazza2019century} and references therein), robotic fingers (and hands) still fall short of their human counterparts. 
Direct-drive fingers are simple to construct, often have high degrees of freedom (3-4 DoF), but are also the largest, and have complicated dynamics due to the apparent inertia of the motors. Tendon-driven hands offer a smaller form factor by moving the motors out of the finger and into the forearm, but other issues arise. A single tendon cannot push and pull a joint, so additional motors are needed in order to maintain fully actuated degrees of freedom, increasing cost and weight. Soft actuated hands, like pneumatic and shape memory, offer a great form factor but lack the degrees of freedom needed for complex manipulation tasks. Although 3-DoF linkage-driven robotic fingers with millimeter-level position repeatability \cite{kim2021integrated, unitreeDex5-1} have been designed, none have demonstrated fingertip trajectory following capability. The key reason is that previous 3-DoF linkage-driven fingers, such as the AIDIN robotics hand, cannot solve the forward kinematics (FK) from the motor commands to the fingertip position analytically. Thus, there is no closed-form expression for the Jacobian, impeding the development of accurate position/velocity controllers for the fingertip motion.
\begin{figure}[t]
    \centering
    \includegraphics[width=0.5\linewidth]{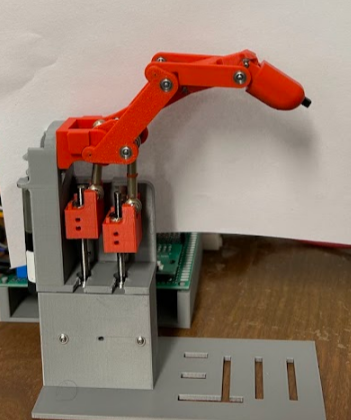}
    \caption{3-DoF Series-Parallel robotic finger prototype with direct-driven abduction and linkage-driven flexion.}
    \label{fig:finger_prototype}
\end{figure}

A recent study~\cite{zawfinger} introduced a linkage-driven, series–parallel robotic finger architecture with an analytic solution to the forward kinematics (FK) and a closed-form expression for the Jacobian. That work demonstrated in simulation that resolved motion rate control (RMRC)~\cite{Whitney1969ResolvedMotion} could achieve desired fingertip trajectories. However, no physical prototype was reported, and the results were limited to simulation-based validation. 
Building on this concept of analytically tractable linkage-driven fingers, we present a physical implementation of a 3-DoF finger (see Figure~\ref{fig:finger_prototype}) with closed-form FK and Jacobian. We further provide an experimental evaluation of task-space trajectory tracking, demonstrating millimeter-level accuracy across a variety of trajectories using RMRC. This is our key contribution, and to the best of our knowledge, this work is one of the first systematic experimental studies of task-space control for a compact, linkage-driven robotic finger.

\section{Related Work}
\label{sec:rw}

As stated earlier, despite its relevance, systematic demonstrations of task-space trajectory tracking in robotic fingers remain scarce, particularly for compact, multi-DoF fingers suitable for integration into anthropomorphic hands.


Direct-drive robotic hands, both commercial (e.g., Allegro~\cite{AllegroHandspecs}, Tesollo~\cite{tessolo5F}, Roboterra X-Hand~\cite{RobotEraXHAND1specs}) and academic (e.g., LEAP~\cite{Shaw2023LEAP}, DLR-HIT~\cite{liu2008DLRHIT}), offer high controllability but suffer from design trade-offs: placing actuators within the fingers increases size, inertia, and control complexity~\cite{kim2021integrated}. As a result, many studies have relied on joint-space control rather than direct task-space trajectory tracking, where performance can be degraded by the dynamics of bulky, high-inertia actuators.

Tendon-driven designs address some of these issues by relocating actuators to the forearm and transmitting forces through tendons, thereby reducing finger size and inertia. However, tendon elasticity, backlash, and friction introduce modeling uncertainty, making precise forward kinematics estimation challenging~\cite{lilge2020FKtendon,anhish2013tendonconrol,Chen2014}. Consequently, much of the existing work remains focused on joint-space control and quasi-static grasping, with few demonstrations of high-precision task-space trajectory tracking.

Linkage-driven hands represent a compelling alternative: they maintain compact finger size comparable to tendon-driven designs but with rigid transmission, ensuring a deterministic mapping between motor positions and fingertip pose. The AIDIN hand~\cite{aidin_robotic_hand}, based on the linkage-driven design in~\cite{kim2021integrated}, demonstrates sub-millimeter repeatability and high force output, but does not provide an analytical forward kinematics model or closed-form Jacobian. Zhang et al.~\cite{Zhang2022ModularRoboticFinger} proposed a modular linkage-driven 2-DoF finger (without abduction) with their study reporting only limited path-following results.

As a result, many researchers continue to adopt simplified grippers like parallel-jaw or 1-DoF fingers, which reduce system complexity but sacrifice dexterity~\cite{wu2015underactuated}. Yet dexterous in-hand manipulation fundamentally requires multiple degrees of freedom per finger and precise task-space control~\cite{Ma2011Dexterity}.

In this work, we address this gap by experimentally characterizing the ability of  a \emph{3-DoF, linkage-driven, series–parallel robotic finger with closed-form kinematics} to follow fingertip trajectories in both flexion/extension and abduction/adduction planes. Our results provide one of the first systematic demonstrations of task-space trajectory tracking for a compact, multi-DoF robotic finger with both abduction/adduction and flexion/extension capability.

\section{Finger Construction and Operation}

\begin{figure}[h]
    \centering
    \includegraphics[width=\linewidth]{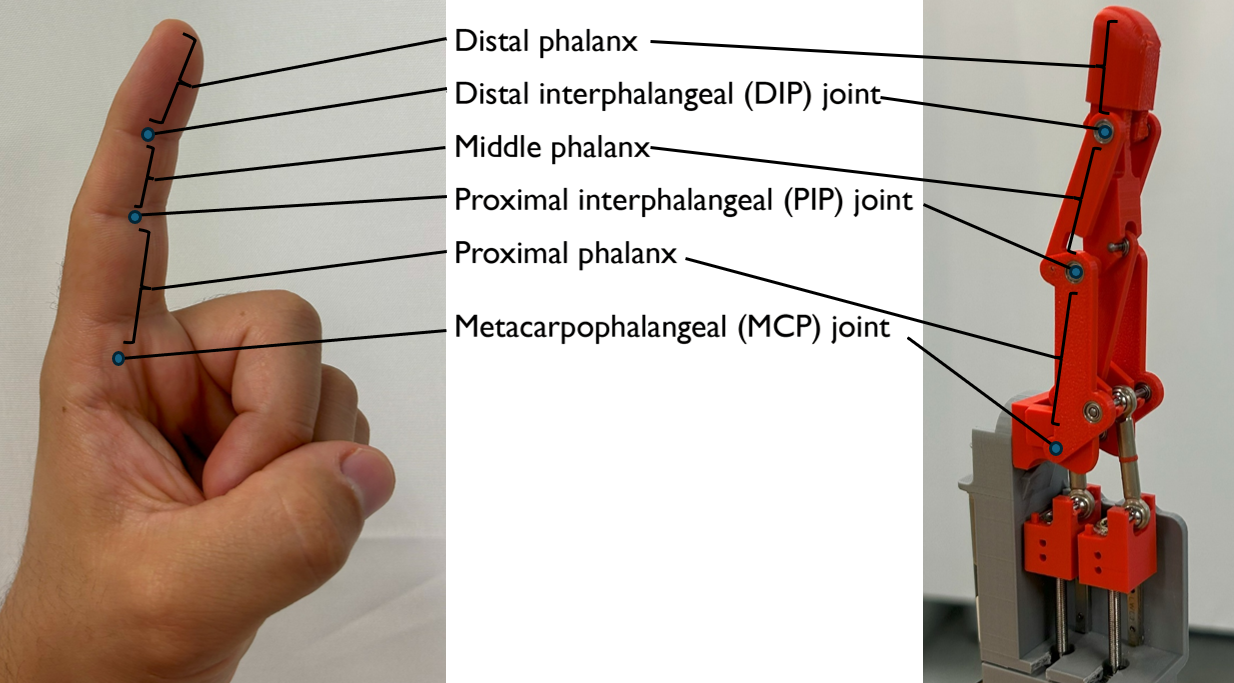}
    \caption{Anatomic comparison of 3-DoF series-parallel finger with a human finger.}
    \label{fig:boneandjointdiagram}
\end{figure}

First, we will define some nomenclature used throughout the paper. The motion of the finger can be described as rotations occurring in two planes: the abduction plane (side-to-side finger 'wagging'), and the flexion plane (finger bending, or curling). Second, the human finger (and our robotic finger) have three degrees of freedom, and are composed of three joints, one universal, and two revolute. Following from Figure \ref{fig:boneandjointdiagram}, we start at the base of the finger with the MCP joint (Metacarpophalangeal), which contributes 1 DoF of rotation in the abduction plane and 1 DoF of rotation in the flexion plane. The middle joint is known as the PIP joint (Proximal Interphalangeal), rotates in the flexion plane, contributing 1 DoF. The top joint is known as the DIP joint (Distal Interphalangeal), and also rotates in the flexion plane, though it is coupled to the PIP joint and therefore does not contribute a DoF. The proximal phalanx connects the MCP to the PIP joint, the middle phalanx connects the PIP joint to the DIP joint, and distal phalanx constitutes the "finger tip" (once again, see Figure \ref{fig:boneandjointdiagram}). 

The term “Series-Parallel” reflects both the structure and the kinematic modeling of the mechanism. The finger consists of four closed loops, which give it a parallel topology. However, by systematically solving these four loop-closure equations, we can extract the joint angles corresponding to the MCP, PIP, and DIP joints, therefore enabling the mechanism to be modeled as a serial manipulator. \cite{zawfinger} used a primitive model, which needed to be adapted according to hardware size constraints, the full process of which is outlined in \cite{baiata2025hybridthumb}. Besides changes to the link dimensions to accommodate hardware, the most notable change to the design is the abduction power transmission. In \cite{zawfinger} all joints are powered by linear actuators with spherical joint rod ends connecting them to the finger joints. We make the simplification of powering the abduction using direct-drive, with miter gears connecting the motor to the joint. Because the axis of rotation is located close to the palm, and does not change location, this change simplifies the kinematics and the design without increasing the size of the mechanism. 

\begin{figure}[h]
    \vspace{5mm}
    \centering
    \includegraphics[width=0.75\linewidth]{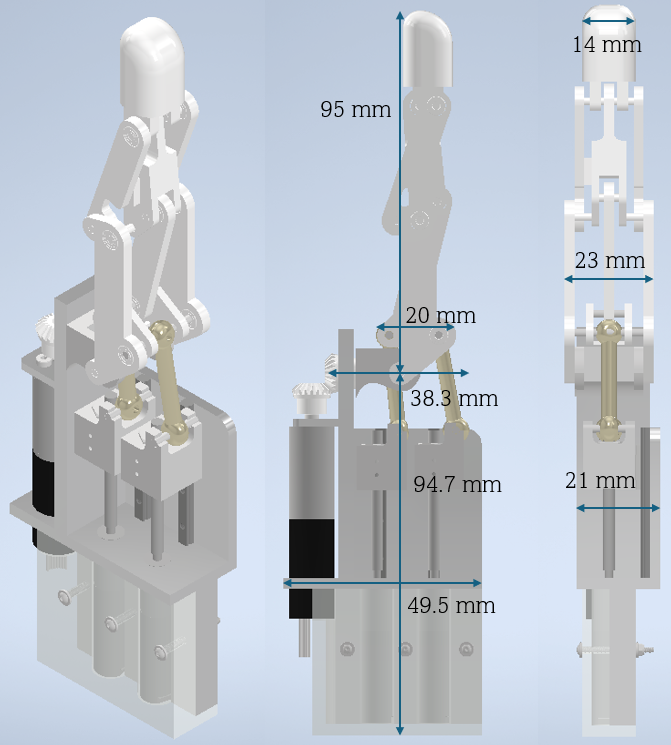}
    \caption{Our 3-DoF series-parallel robotic finger dimensions ($23 \times 20 \times95$ mm) is comparable in size to the AIDIN Robotics finger ($20 \times 21.5 \times 101$ mm) and Shadow Robotics Finger ($20 \times 18 \times 100$ mm)}
    \label{fig:fingCAD}
\end{figure}

To power the two flexion DoF, Faulhaber 0824-BLDC motors were used in combination with Faulhaber 08L-SL lead screws. The abduction is powered by a Faulhaber 1226-BLDC motor. The finger takes roughly two hours to put together, excluding print times, and weighs 105.7 grams. The dimensions seen in Figure \ref{fig:fingCAD} show the finger is comparable in size to the AIDIN robotics \cite{aidin_robotic_hand} finger, and the Shadow Robotics \cite{ShadowHand2024techreport} finger.

\subsection{Controller/Planner Design}
To present the controller algorithms, we will first need some notation and definitions.
Let $ \texttt{FK}(\mathbf{M}) \;\mapsto\; \mathbf{X}$ be the forward kinematics map, $\texttt{IK}(\mathbf{X}) \;\mapsto\; \mathbf{M}$ be the inverse kinematics map,  and $\texttt{J}(\mathbf{M}) \;\mapsto\; \mathbf{J} = \frac{\partial \mathbf{X}}{\partial \mathbf{M}}$ be the motor position dependent Jacobian map. Here $\mathbf{M} \in \mathbb{R}^3$ are the motor positions and 
$\mathbf{X} \in \mathbb{R}^3$ are the Cartesian coordinates of the end effector relative to the world frame at the MCP joint. The Jacobian $\mathbf{J} \in \mathbb{R}^{3 \times 3}$ relates the fingertip velocity to the motor velocities. The derivation of the forward and inverse kinematics, as well as the Jacobian, are based on~\cite{zawfinger}, and can be found in the Supplementary Materials. Note that in both the forward and inverse kinematics, the joint angles $\mathbf{Q} = \{q_1, q_2, q_3, q_4, \beta\} \in \mathbb{R}^5$ are also calculated as an intermediate step using the maps $ \texttt{M2Q}(\mathbf{M}) \;\mapsto\; \mathbf{Q}$ and $\texttt{X2Q}(\mathbf{X}) \;\mapsto\; \mathbf{Q}$. Additionally, if the joint angles are known, we can find the motor positions using the map $\texttt{Q2M}(\mathbf{Q}) \;\mapsto\; \mathbf{M}$.

Any low-level motor controller functions should be self-evident by name, i.e. \texttt{ReadEncoders()} reads the motor encoders. The low-level motor control (both positioning and velocity) was done using Faulhaber's MC3001P motor controller. An Arduino Uno R4 was used for the high level motion control. The Waveshare SN65HVD230 CAN Transceiver was used to communicate between the Arduino and the three Faulhaber controllers. 

\subsubsection{Joint Space Planner}

Algorithm~\ref{algo1} gives the pseudocode for the joint space controller. This controller takes in a desired joint configuration, $\mathbf{Q}_\text{goal}$, and uses linear interpolation in the joint space to iteratively compute a sequence of motor position commands to reach $\mathbf{Q}_\text{goal}$ (lines $7$ to $11$). A key aspect here is that since the joints are not directly driven, we need to ensure that the planned joint space motion does not lead to motor commands that exceed the maximum speed limit of the motors. We use a simple procedure to implement this where we guess an initial time $t$ to reach the goal using the maximum motor speeds (line $4$ of Algorithm~\ref{algo1}), update $t$ to $t+\delta$ if the maximum motor speed is exceeded and continue until we get feasible motor commands for reaching the goal (lines $12$ to $14$). We chose $\delta = 0.25$ for our experiments. As in the direct drive case, this joint space-based controller is robust to singularities and always returns a valid motion, so long as $\mathbf{Q}_\text{goal}$ is within the joint limits.

While the proposed controller guarantees feasibility by iteratively increasing $t$ until the motor speed constraints are satisfied, this procedure can be computationally inefficient. 
In ongoing work, we are developing a method to directly compute the minimum feasible $t$ that satisfies all motor velocity limits, thereby eliminating the need for iterative adjustment and improving the controller’s execution speed.


\begin{algorithm}[H]
\caption{Single Point Joint Space Planner} \label{algo1}
\begin{algorithmic}[1]
\Require $\mathbf{Q}_{\text{goal}}$, dt $= 0.01$ s, per-motor max speeds $\mathbf{v}_{\max}$
\State $\mathbf{M}_{\text{cur}} \gets \texttt{ReadEncoders}()$
\State $\mathbf{Q}_{\text{cur}} \gets \texttt{M2Q}(\mathbf{M}_{\text{cur}})$
\State $\Delta \mathbf{M} \gets \mathbf{M}_{\text{goal}} - \mathbf{M}_{\text{cur}}$
\State $t \gets \dfrac{\max_i |\Delta \mathbf{M}_i|}{v_{\max}}$ \Comment{initial time guess}
\State Initialize list $\mathcal{M} \gets [\ ]$ \Comment{trajectory in motor space}
\Repeat
    \State $n = ~\lceil \frac{t}{dt} \rceil$
    \For{$k = 0 ~\text{to} ~n$}
        \State $\mathbf{Q}_k = \mathbf{Q}_\text{cur} + \frac{k}{n}(\mathbf{Q}_\text{goal}-\mathbf{Q}_\text{cur})$ 
        \State $\mathcal{M}[k] \gets \texttt{Q2M}(\mathbf{Q}_k)$
    \EndFor
    \If{$\max_k\left\{\dfrac{\mathcal{M}[k]-\mathcal{M}[k-1]}{\text{dt}}\right\} > v_{\max}$}
        \State $t \gets t + \delta $ \Comment{increase duration and try again}
    \EndIf
\Until{no speed limit is exceeded}
\State \texttt{CommandMotorPositions}($\mathcal{M}$)
\end{algorithmic}
\end{algorithm}

\subsubsection{Task Space Planner}

Algorithm \ref{algo2} shows the task space planner. The task space planner allows the user to input up to N points, where N is determined by hardware memory limits (for our Arduino Uno we allow 50 points to be input). We assume that all points in the path are within the finger's workspace, with no singularities between or at points. In case there is a singularity between the last position of the finger and the first point of the input path, the joint space planner moves the finger to the first point in the path. The forward kinematics are used to find the current task space position, as well as the Jacobian. The difference between the current position and the next goal point is calculated. This vector, $\mathbf{E}$, represents the error between our current and goal location. The unit vector of $\mathbf{E}$, $\hat{\mathbf{X}}_\text{cur}$ represents the direction the fingertip must move in order to reach the goal location. Multiplying $\hat{\mathbf{X}}_\text{cur}$ by $v_\text{desired}$ gives the desired task space velocity vector. Multiplying this by the inverse of the Jacobian gives the necessary motor velocities, $\dot{\mathbf{M}}$. If any of the necessary motor velocities exceed our max motor speeds, all velocities must be scaled down proportionally, so the direction of the task space velocity vector remains the same. This loop continues until the error in each Cartesian direction is less than 0.1 mm, where the planner then moves on to the next point, until all points are complete.

Note that, when a desired point in the task space is specified either the joint space planner OR the task space planner can be used to reach it. The joint space planner should be used if the path is of no consequence, as a solution is guaranteed. The task space planner should be used if the fingertip should follow a straight line path to reach the point.

\begin{algorithm}
\caption{Task Space Planner} \label{algo2}
\begin{algorithmic}[1]
    \Require $\mathbf{X}_\text{list} \in \mathbb{R}^{N \times 3}, \; $, desired task space speed $\text{v}_\text{desired} $, per-motor max speeds $\mathbf{v}_{\max}$
    \State Send $\mathbf{X}_\text{list}[1]$ (first point) to Joint Space Planner
    \State $\mathbf{M}_{\text{cur}} \gets \texttt{ReadEncoders}()$
    \State $\mathbf{X}_{\text{cur}} \gets \texttt{FK}(\mathbf{M}_{\text{cur}})$
    \For {remaining points in $\mathbf{X}_\text{list}$}
        \State Set $ \mathbf{E} = [\text{Error}_x, \text{Error}_y, \text{Error}_z] = [100,100,100]$ mm  
        \Repeat
        \State $\mathbf{M}_{\text{cur}} \gets \texttt{ReadEncoders}()$
        \State $\mathbf{X}_{\text{cur}} \gets \texttt{FK}(\mathbf{M}_{\text{cur}})$
        \State $\mathbf{E} = \mathbf{X}_{\text{list}}[k]-\mathbf{X}_{\text{cur}}$ \Comment{current error}
        \State $\hat{\mathbf{X}}_\text{cur} \gets \frac{\mathbf{E}}{\lVert\mathbf{E}\rVert}$ \Comment{unit vector pointing to desired location}
        \State $\textbf{J} \gets \texttt{J}(\mathbf{M}_{\text{cur}})$ 
        \State $\dot{\mathbf{X}} \gets \hat{\mathbf{X}}_\text{cur}\text{v}_\text{desired}$
        \State $\dot{\mathbf{M}} \gets \mathbf{J}^{-1}\dot{\mathbf{X}}$
        \State Check if $\dot{\mathbf{M}} \leq \mathbf{v}_{\max}$, scale down accordingly if not
        \State \texttt{CommandMotorVelocities}($\dot{\mathbf{M}}$)
        \Until{All errors are $\leq 0.1$ mm}
        \State Iterate to next point
    \EndFor
    \State \texttt{CommandMotorVelocities}($\mathbf{0}$)
\end{algorithmic}
\end{algorithm}

\section{Results}
\label{sec:results}
Six different trajectories will be presented in this section, three in the flexion plane and three in the abduction plane. The same three paths will be presented in each plane: a square, a circle, and a step path. The square path shows the finger is capable of following long, straight lines. The circle path demonstrates the ability to approximate curves using many smaller lines. The step path shows that very precise path following is possible, even when following a very small, rapidly changing path. We separated the paths into two 2-D planes in order to increase our measuring accuracy. 

\subsection{Measuring Method}\label{measuring}
Accurately tracking a point in 3-D space at a low cost is difficult. Commercial motion-capture systems such as the OptiTrack PrimeX41 \cite{OptiTrack_PrimeX41_2025} claim sub-millimeter precision but are prohibitively expensive. Open-source alternatives like AprilTags \cite{wang2016apriltag} are designed for real-time tracking of larger objects and are less suited for our application, which requires offline tracking of fine fingertip trajectories. We opted to use a pixel-to-metric technique to track the fingertip motion. 

We attached a 3 × 3 × 1 mm rectangular prism to the fingertip, and tracked it with Segment Anything Model 2 (SAM 2) \cite{ravi2024sam2}. Because the paths were traced near the image center using a relatively flat object, lens distortion was neglected. The script presents the user the first frame and prompts them to click on the prism. SAM 2 generates a mask of the prism, and the centroid of the mask is calculated. The pixel coordinates of the centroid are then used as the click prompt to generate the mask in the next frame, looping until the full video is processed. For the abduction paths the centroid coordinates were extracted and plotted to mark the fingertip location. For the flexion paths, the mask was estimated as a square and the midpoint of the left-most edge was used as the fingertip location. By manually measuring the pixel length of one side of the prism, we can create our pixel to millimeter (P/mm) ratio and convert the pixel coordinates of the fingertip to millimeters. For each path, the finger was manually calibrated to the starting position using a 3-D printed structure.

Uncertainty arising from the prism size, P/mm ratio, and fingertip localization was propagated for each point according to \cite{bevington2002data}. The uncertainty in the prism size comes from the caliper manufacturer's specifications, while the uncertainty in the P/mm ratio and fingertip localization had to be manually estimated. For the P/mm ratio, the prism was black and the background was white, giving a 2-5 pixel gray blur around the prism. For the fingertip localization, random frames of each video were sampled and the difference between the actual fingertip location and the one generated by SAM 2 were measured. This difference was typically only 1-3 pixels. 

\subsection{Fingertip Trajectory Following Results}

The error we wish to quantify is the \textit{trajectory norm error}, defined as the norm of the deviation between measured and expected trajectory positions. This means we need to calculate where the finger should be, and actually was, at each time-step throughout the trajectory. This can be done by assuming the video capture of the demonstrations used a constant frame rate (30 $fps$). As explained in Section \ref{measuring}, each data point (position in space) is one frame apart, so each point's index represents the number of frames, $F$, that have passed since the start of the demonstration. The expected trajectory can be calculated by dividing the desired speed, $v_\text{desired}$, by the frame rate, 30 $fps$, giving the expected movement per frame $s$. By interpolating along the input path at step $s$, we can find the expected position at each frame $F$. Although computing full path uncertainty is nontrivial, we evaluate performance using the maximum trajectory error and its uncertainty, and report average maximum trajectory errors across trials via inverse-variance weighting.

\begin{figure}[h]
    \centering
    \begin{minipage}{0.48\linewidth}
        \centering
        \includegraphics[width=\linewidth]{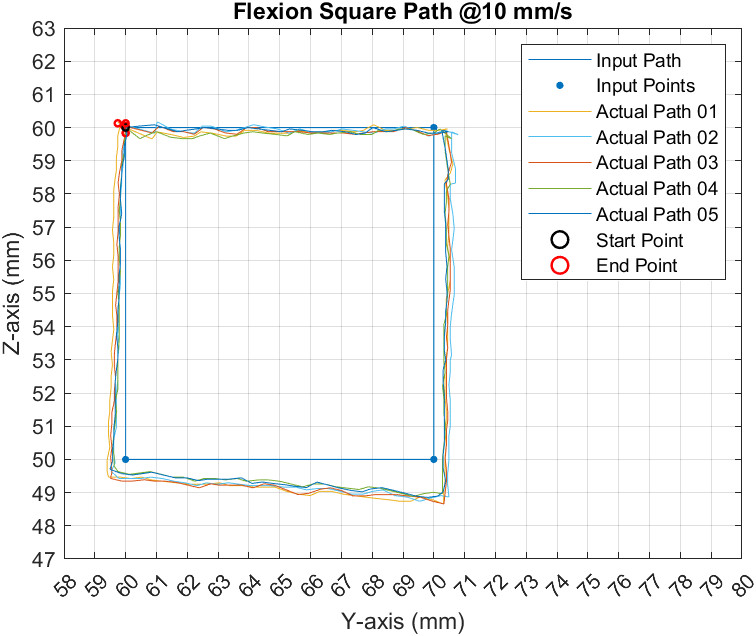}
        \caption{Square Path in the Flexion Plane at 10 mm/s}
        \label{fig:square10all}
    \end{minipage}\hfill
    \begin{minipage}{0.48\linewidth}
        \centering
        \includegraphics[width=\linewidth]{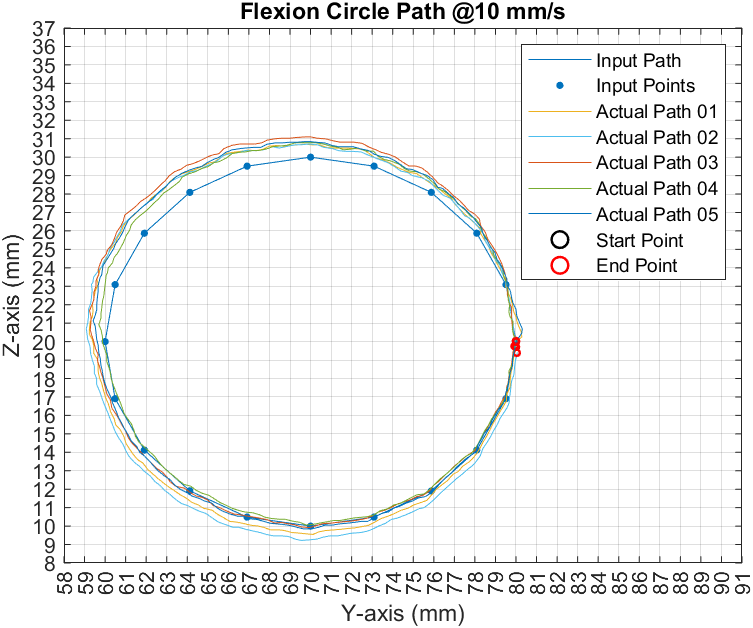}
        \caption{Circle Path in the Flexion Plane at 10 mm/s}
        \label{fig:circle10all}
    \end{minipage}
\end{figure}

\begin{figure}[h]
    \centering
    \begin{minipage}{0.48\linewidth}
        \centering
        \includegraphics[width=\linewidth]{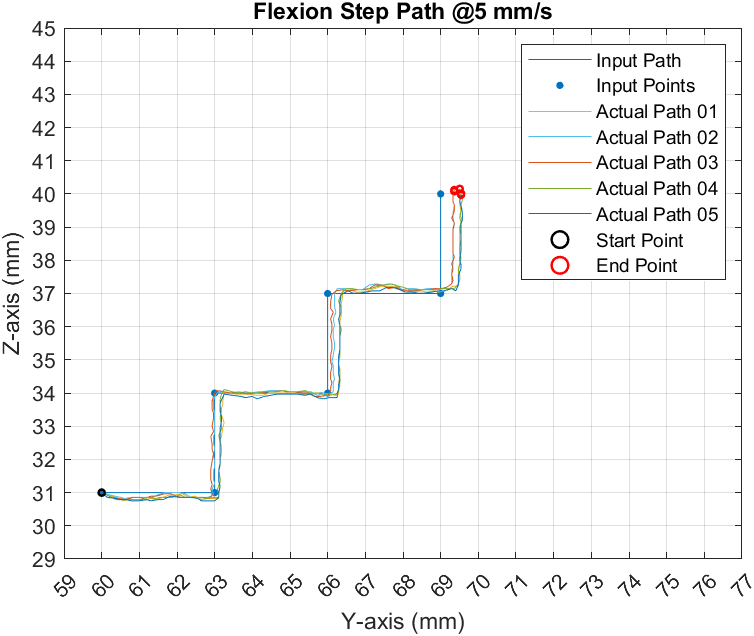}
        \caption{Step Path in the Flexion Plane at 5 mm/s}
        \label{fig:step5all}
    \end{minipage}\hfill
    \begin{minipage}{0.48\linewidth}
        \centering
        \includegraphics[width=\linewidth]{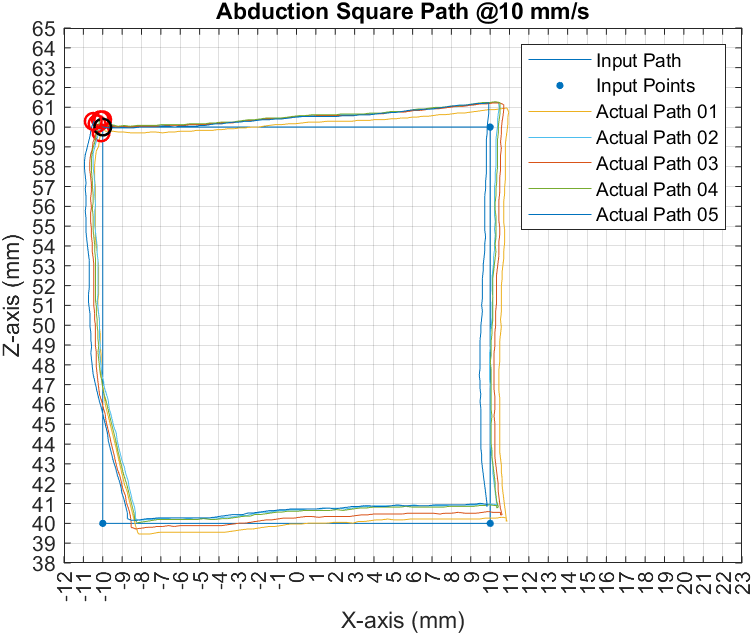}
        \caption{Square Path in the Abduction Plane at 10 mm/s}
        \label{fig:adsquare10all}
    \end{minipage}
\end{figure}

\begin{figure}[H]
    \centering
    \begin{minipage}{0.48\linewidth}
        \centering
        \includegraphics[width=\linewidth]{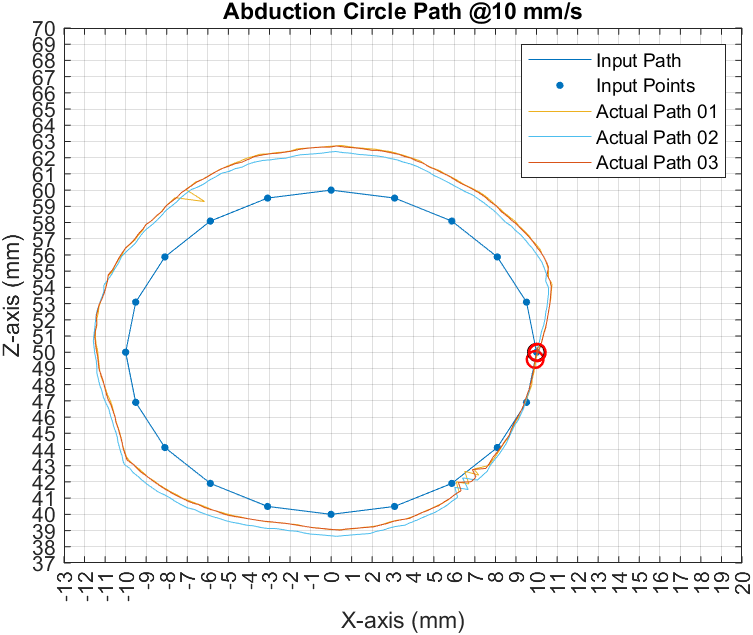}
        \caption{Circle Path in the Abduction Plane at 10 mm/s}
        \label{fig:adcircle10all}
    \end{minipage}\hfill
    \begin{minipage}{0.48\linewidth}
        \centering
        \includegraphics[width=\linewidth]{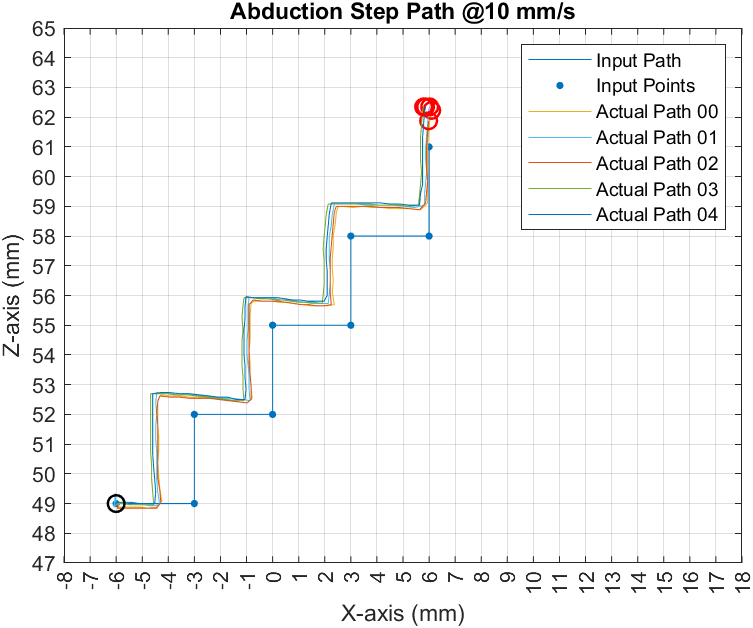}
        \caption{Step Path in the Abduction Plane at 10 mm/s}
        \label{fig:adstep10all}
    \end{minipage}
\end{figure}

First, we will examine the flexion plane results. Five trials for each trajectory were performed, with the square and circle paths done at 10 mm/s, and the step path at 5 mm/s. Figures \ref{fig:square10all}, \ref{fig:circle10all}, and \ref{fig:step5all} show each trajectory's five measured trials. All abduction plane paths were performed at 10 mm/s. Five trials were completed for the square and step paths, but only 3 could be completed for the circle path (shadows cast by the fingertip made it difficult to capture the mask). All trajectory following results can be found in Table \ref{tab:traj_results}. Despite the abduction plane error being significantly larger than the flexion plane error, the results seen in Figures \ref{fig:adsquare10all}, \ref{fig:adcircle10all}, and \ref{fig:adstep10all} appear not to deviate that far from the paths. This suggests the issue lies in the velocity following, and not the path.

\begin{table}[H]
\centering
\caption{Maximum norm error $\bar{x}$ and uncertainty $\sigma_{\bar{x}}$ for trajectory following in the flexion and abduction planes.}
\label{tab:traj_results}
\begin{tabular}{lcccc}
\toprule
\multirow{2}{*}{Path} & \multicolumn{2}{c}{Flexion Plane} & \multicolumn{2}{c}{Abduction Plane} \\
\cmidrule(lr){2-3} \cmidrule(lr){4-5}
 & $\bar{x}$ (mm) & $\sigma_{\bar{x}}$ (mm) & $\bar{x}$ (mm) & $\sigma_{\bar{x}}$ (mm) \\
\midrule
Square & 1.46 & 0.24 & 6.44 & 0.31 \\
Circle & 1.75 & 0.24 & 5.77 & 0.20 \\
Step   & 0.58 & 0.13 & 2.51 & 0.13 \\
\bottomrule
\end{tabular}
\end{table}

\subsection{Fingertip Path Following Results}

We can analyze the path following capabilities by modifying the desired speed to that of the measured speed. This means the new $s$ is simply the path length divided by the total number of frames for a given demonstration. We can see in Table \ref{tab:path_results} the flexion plane results do not change, indicating high accuracy in the velocity following. The abduction plane error decreases for every path, confirming the issue lies in the velocity following and not the path following. The increased error in both abduction trajectory and path following may stem from a lack of an encoder on the abduction motor. Relying on just the analog hall sensors, the motor could not be tuned as accurately as the flexion motors, which have 4096-bit incremental encoders. Backlash in the 3-D printed miter gears may also contribute to this error. 

\begin{table}[H]
\centering
\caption{Maximum norm error $\bar{x}$ and uncertainty $\sigma_{\bar{x}}$ for path following in the flexion and abduction planes} 
\label{tab:path_results}
\begin{tabular}{lcccc}
\toprule
\multirow{2}{*}{Path} & \multicolumn{2}{c}{Flexion Plane} & \multicolumn{2}{c}{Abduction Plane} \\
\cmidrule(lr){2-3} \cmidrule(lr){4-5}
 & $\bar{x}$ (mm) & $\sigma_{\bar{x}}$ (mm) & $\bar{x}$ (mm) & $\sigma_{\bar{x}}$ (mm) \\
\midrule
Square & 1.46 & 0.24 & 2.74 & 0.30 \\
Circle & 1.75 & 0.24 & 2.80 & 0.36 \\
Step   & 0.58 & 0.13 & 2.01 & 0.07 \\
\bottomrule
\end{tabular}
\end{table}

An additional metric we evaluate is the difference between the start and end positions for the square and circle paths. For these paths the start and end points are the same input, essentially testing repeatability of the task planner to reach a given point. These results (see Table \ref{tab:start_end_norm}), along with the high repeatability of error location between successive paths of a certain trajectory, lead us to believe much of the error comes from poor calibration and inconsistencies between the actual link lengths and those used in the kinematic controller. 

\begin{table}[H]
\centering
\caption{Norm $\bar{d}$ and uncertainty $\sigma_{\bar{d}}$ of start and end points for the square and circle paths in the flexion and abduction planes}
\label{tab:start_end_norm}
\begin{tabular}{lcccc}
\toprule
\multirow{2}{*}{Path} & \multicolumn{2}{c}{Flexion Plane} & \multicolumn{2}{c}{Abduction Plane} \\
\cmidrule(lr){2-3} \cmidrule(lr){4-5}
 & $\bar{d}$ (mm) & $\sigma_{\bar{d}}$ (mm) & $\bar{d}$ (mm) & $\sigma_{\bar{d}}$ (mm) \\
\midrule
Square & 0.13 & 0.01 & 0.39 & 0.01 \\
Circle & 0.18 & 0.01 & 0.18 & 0.01 \\
\bottomrule
\end{tabular}
\end{table} 

\subsection{Further Validation of Tracking Error}

\begin{figure}[H]
    \centering
    \includegraphics[width=.65\linewidth]{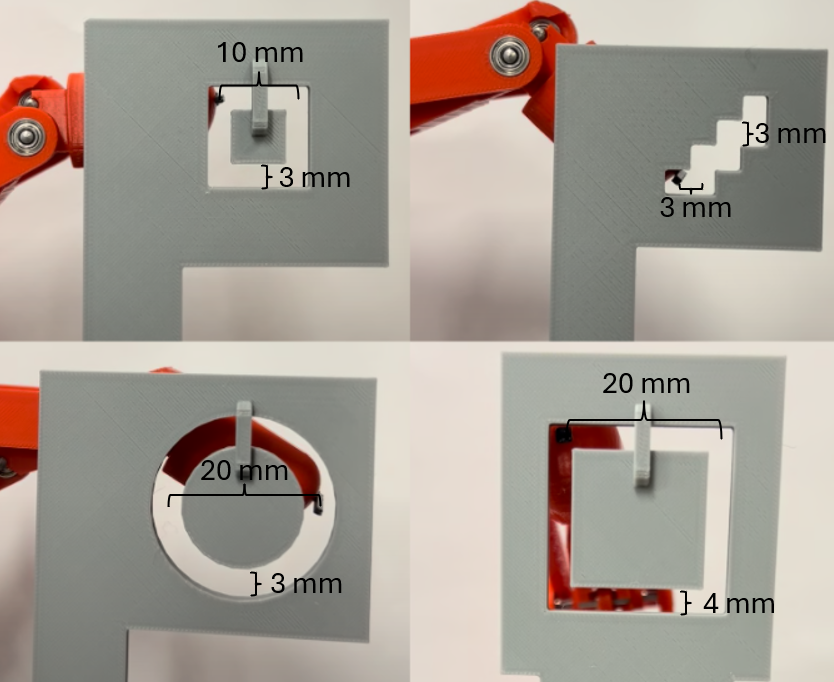}
    \caption{3-D Printed Path Following Structures}
    \label{fig:peg_and_path}
\end{figure}

There is a relatively high uncertainty in the error estimation.  For example, looking at the flexion square results, a 95\% confidence interval will give $\pm0.48$ mm, which is nearly 33\% of the measured error. A simple way to validate the accuracy of our measurements is to use a physical structure of the desired path. A peg is attached to the end of the fingertip as it traces a path in a 3-D printed fixture. As long as the peg does not contact the fixture, you can determine the upper bound of possible error. First, starting with larger paths and then working our way down, we were able to achieve good results with 3 mm wide flexion paths ($\pm$1.5 mm on each side of the path), and 4 mm wide abduction paths ($\pm$2.0 mm on each side). Figure \ref{fig:peg_and_path} shows the four 3-D printed path fixtures with the finger at the starting position in each. Videos of the SAM 2 and 'peg in path' demonstrations will be attached.

In all demonstrations the finger was able to move through the path without hitting the 3-D printed fixture. The path boundaries also act as visual reference, making it easier to discern the motion of the fingertip with the human eye. For example, in the flexion plane square path, you can see the finger get closer to the boundary in the bottom right corner, matching the measured data. There also appears to be some small vibrations in the fingertip motion, which we attribute to the build quality. Over time, the integrity of the press-fit joints in the 3D-printed components degrades, leading to increased mechanical play in the fingertip. 


\section{Summary and Future Work}

In this work, we presented the experimental characterization of a three-degree-of-freedom, linkage-driven, series–parallel robotic finger capable of precise task-space trajectory tracking. Our results demonstrate that accurate fingertip motion control is achievable even with a prototype fabricated from 3D-printed parts, highlighting the promise of linkage-driven architectures for dexterous manipulation. Looking forward, we aim to conduct a more rigorous quantification of trajectory-tracking performance by reporting error distributions and uncertainty bounds in full 3D space. From a hardware perspective, we plan to improve the build quality by transitioning from 3D-printed components to machined aluminum links and replacing lead screws with ball screws. These upgrades will reduce mechanical play, enable back-drivability, and further enhance positioning accuracy. With these improvements in place, we will proceed to experimentally quantify the finger’s force resolution and control bandwidth. Finally, we will extend this work to the construction of a complete robotic hand composed of multiple fingers following the presented design and a four-degree-of-freedom thumb designed using the same principles. This will allow us to experimentally evaluate dexterous, multi-fingered in-hand manipulation tasks under task-space control, paving the way toward a fully functional, high-performance anthropomorphic robotic hand.



\bibliographystyle{IEEEtran} 
\bibliography{main}

\section{Supplementary Materials}

\subsection{Deriving the Loop-Closure Equations}\label{sec:deriving_loops}

\begin{figure}[H]
    \centering
    \includegraphics[width=.7\linewidth]{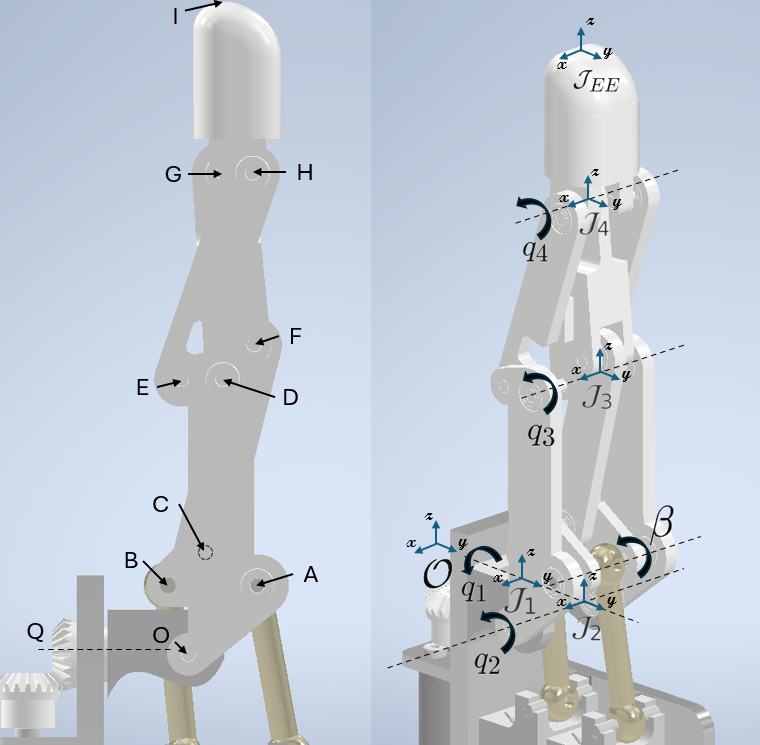}
    \caption{Finger Joint and Linkage Diagram}
    \label{fig:jointvis}
\end{figure}

In order to analyze the finger's kinematics we assign a letter to each joint, allowing us to describe the vectors which construct the finger. Additionally, reference frames $\mathcal{J}_n$ for n = 1,2,3,4, are placed at the "serial" joints, and $\mathcal{O}$ is the base frame. We name them "serial" joints as the rotations about these joints will be used to construct the transformation matrices which describe the forward, inverse, and differential kinematics (as you would with a traditional serial manipulator). The motion of these serial joints also correspond to the same motions found in the human finger. $q_1$, which represents the rotation around the y-axis of $\mathcal{J}_1$, is the abduction angle (side-to-side motion) of the MCP joint (bottommost joint). $q_2$, which represents the rotation around the x-axis of $\mathcal{J}_2$, is the flexion angle (curling motion) of the MCP joint. $q_3$, which represents the rotation around the x-axis of $\mathcal{J}_3$, is the flexion angle of the IP joint (middle joint). All three of these rotations are independent and contribute degrees of freedom. $q_4$, represents the rotation around the x-axis of $\mathcal{J}_4$, and is the flexion angle of the DIP (topmost joint). Unlike the other joints, $q_4$ is coupled to the rotation of $q_3$, just like in the human finger. Finally there is $\beta$, which describes rotation of the bell crank about the $A$ joint. Though not used directly in the kinematic transformation matrices, finding $\beta$ is necessary in finding $q_3$ and $q_4$. Although $\mathcal{O}$, $\mathcal{J}_1$, and $\mathcal{J}_2$ coincide spatially, they have been separated in Figure \ref{fig:jointvis} for clarity. The motor positions will be represented by $m_n$, where $m_1$ is in radians, and $m_2$ and $m_3$ are in millimeters. While $m_2$ and $m_3$ are scalars that exist purely in the z-plane, $\textbf{M}_2$ and $\textbf{M}_3$ represent the vectorized versions.

There are four loop-closure equations that describe the parallel structure of the mechanism. Each loop consists of a known scalar $l_n$, whose length we can equate to the magnitude of the known vectors and some unknown rotation which is embedded in a rotation matrix. $C$ and $S$ represent the cosine and sine functions, respectively.

The only direct-drive joint is $q_1$, as such it is directly proportional to $m_1$, giving us the first rotation matrix:

\begin{equation}
\textbf{R}_{q_1} = \begin{bmatrix}
Cq_1 & 0 & Sq_1 \\
0 & 1 & 0\\
-Sq_1 & 0 & Cq_1
\end{bmatrix} = \begin{bmatrix}
Cm_1 & 0 & Sm_1 \\
0 & 1 & 0\\
-Sm_1 & 0 & Cm_1
\end{bmatrix}
\end{equation}
The first loop-closure equation represents the translation of motion from $m_2$ to $q_2$:
\begin{equation}
    \|{l}_2\| = \|{\textbf{P}_2 + \textbf{M}_2 -\textbf{R}_2\vec{OA}} \| 
    \label{2.4}
\end{equation}
where $\textbf{P}_2$ is the vector from $O$ to $\textbf{M}_2$ at the zero position, $l_2$ is the length of the rod between $\textbf{M}_2$ and $A$, and $\textbf{R}_2$ is the rotation matrix: 
\begin{equation}
\textbf{R}_{2} = \begin{bmatrix}
Cq_1 & Sq_1Sq_2 & Sq_1Cq_2 \\
0 & Cq_2 & -Sq_2 \\
-Sq_1 & Cq_1Sq_2 & Cq_1Cq_2
\end{bmatrix}
\end{equation}
Similarly, $\beta$ is found from $m_3$ using:
\begin{equation}
    \|\textbf{l}_3\| = \|\textbf{P}_3 + \textbf{M}_3 - \textbf{R}_2\vec{OA} - \textbf{R}_3\vec{AB} \|
    \label{2.7}
\end{equation}
where $\textbf{P}_3$ is the vector from $O$ to $\textbf{M}_3$ at the zero position, $l_3$ is the length of the rod between $\textbf{M}_3$ and $B$, and $\textbf{R}_3$ is the rotation matrix: 
\begin{equation}
\textbf{R}_3 = \begin{bmatrix}
Cq_1 & Sq_1S(q_2+\beta_1) & Sq_1C(q_2+\beta_1) \\
0 & C(q_2+\beta_1) & -S(q_2+\beta_1) \\
-Sq_1 & Cq_1S(q_2+\beta_1) & Cq_1C(q_2+\beta_1)
\end{bmatrix}
\end{equation}
Now that we have $\beta$, we can find $q_3$ using,
\begin{equation}
     \|\textbf{l}_4\| = \|-\textbf{R}_{\beta_1}\vec{AC} + \vec{AD} + \textbf{R}_{q_3}\vec{DF}\|
    \label{2.12}
\end{equation}
where $l_4$ is the linkage between joints $F$ and $C$, and $\textbf{R}_{\beta_1}$ and $\textbf{R}_{q_3}$ are the rotation matrices
\begin{equation}
    \textbf{R}_{\beta_1} = \begin{bmatrix}
1 & 0 & 0 \\
0 & C\beta_1 & -S\beta_1 \\
0 & S\beta_1 & C\beta_1        
    \end{bmatrix}
\end{equation}
\begin{equation}
    \textbf{R}_{q_3} = \begin{bmatrix}
1 & 0 & 0 \\
0 & Cq_3 & -Sq_3 \\
0 & Sq_3 & Cq_3        
    \end{bmatrix}
\end{equation}
Note how $q_3$ has no dependence on $q_2$, i.e. the middle joint (and therefore top joint, as they are coupled), can flex/extend independently of the bottom joint's flexion/extension. The fourth and final loop-closure equation is, 
\begin{equation}
   \|l_5\| =  \| \textbf{R}_{q_3}\vec{DG} + \textbf{R}_{q_3}\textbf{R}_{q_4}\vec{GH} - \vec{DE}\|
    \label{2.16}
\end{equation}
where $l_5$ is the distance between $E$ and $H$, and $\textbf{R}_{q_4}$ is 
\begin{equation}
    \textbf{R}_{q_4} = \begin{bmatrix}
1 & 0 & 0 \\
0 & Cq_4 & -Sq_4 \\
0 & Sq_4 & Cq_4        
    \end{bmatrix}
\end{equation}

\subsection{Solving the Loop-Closure Equations}

As previously discussed, each loop-closure equation consists of a scalar ($l_n$), various vectors belonging to $\mathbb{R}^3$, some of which are pre-multiplied by rotation matrices in $\mathrm{SO}(3)$, along with the unknown angle. The known vectors and matrices can be simplified to a single vector, which we will call $\textbf{U}_n$. The unknown rotation and accompanying vector can be denoted $\textbf{R}_n\textbf{V}_n$. By definition of the loop-closure equations, we know the magnitude of $\textbf{U}_n$ and the unknown rotation, $\textbf{R}_n\textbf{V}_n$, must be equal to $l_n$, i.e. 
\begin{equation}
    l_n^2 = (\textbf{U}_n-\textbf{R}_n\textbf{V}_n)^T(\textbf{U}_n-\textbf{R}_n\textbf{V}_n)
    \label{2.17}
\end{equation}
Remembering that $\textbf{R}^T\textbf{R}=\textbf{I}$ we can expand the above equation into:
\begin{equation}
    {l}_n^2 = \|\textbf{U}_n\|^2 + \|\textbf{V}_n\|^2 + 2\textbf{U}_n^T\textbf{R}_n\textbf{V}_n
\end{equation}
Rearranging the known terms to the right-hand side, we get:
\begin{equation}
    \textbf{U}_n^T\textbf{R}_n\textbf{V}_n =  \frac{\|\textbf{U}_n\|^2 + \|\textbf{V}_n\|^2 - {l}_n^2}{2} 
\end{equation}
The right-hand side of this equation will be denoted by $\text{S}_n$. Multiplying out the left-hand side yields a long polynomial that can then be simplified to the form: 
\begin{equation}
   \text{A}_n \sin(q_n) +\text{B}_n \cos(q_n) +\text{C}_n = 0
\end{equation}
where $\text{C}_n$ includes $\text{S}_n$, along with any other constants resulting from the multiplication of the left-hand side. The solution to this equation is a known trigonometric problem and is given here:
\begin{equation}
    q_n = \arctan2(\text{A}_n, \text{B}_n) \pm \arccos\left(\frac{-\text{C}_n}{\sqrt{\text{A}_n^2 + \text{B}_n^2}}\right)
\end{equation}

\subsection{Constructing the Transformation Matrices}

Using the joint rotations of $q_1$, $q_2$, $q_3$, $q_4$, and the appropriate link lengths, we can treat the finger as a serial manipulator and construct homogeneous transformation matrices. This will allow us to calculate the end-effector location for a given set of motor inputs. However, we still need to introduce a frame at the fingertip, $\mathcal{J}_{EE}$. 

\begin{equation*}
    \textit{T}_{\mathcal{O}}^{\mathcal{J}_1} =
    \begin{bmatrix}
Cq_1 & 0 & Sq_1 & 0\\
0 & 1 & 0 & 0\\
-Sq_1 & 0 & Cq_1 & 0 \\
0 & 0 & 0 & 1
\end{bmatrix},     
\textit{T}_{\mathcal{J}_1}^{\mathcal{J}_2} =
    \begin{bmatrix}
1 & 0 & 0 & 0\\
0 & Cq_2 & -Sq_2 & 0\\
0 & Sq_2 & Cq_2 & 0 \\
0 & 0 & 0 & 1
\end{bmatrix}
\end{equation*}

\begin{equation*}
    \textit{T}_{\mathcal{J}_2}^{\mathcal{J}_3} =
    \begin{bmatrix}
1 & 0 & 0 & 0\\
0 & Cq_3 & -Sq_3 & \text{OD}_y\\
0 & Sq_3 & Cq_3 & \text{OD}_z \\
0 & 0 & 0 & 1
\end{bmatrix}
\end{equation*}
\begin{equation*}
    \textit{T}_{\mathcal{J}_3}^{\mathcal{J}_4} =
    \begin{bmatrix}
1 & 0 & 0 & 0\\
0 & Cq_4 & -Sq_4 & \text{DG}_y\\
0 & Sq_4 & Cq_4 & \text{DG}_z \\
0 & 0 & 0 & 1
\end{bmatrix},
    \textit{T}_{\mathcal{J}_4}^{\mathcal{J}_{EE}} =
    \begin{bmatrix}
1 & 0 & 0 & 0\\
0 & 1 & 0 & 0\\
0 & 0 & 1 & \text{GI}_z \\
0 & 0 & 0 & 1
\end{bmatrix}
\end{equation*}
\begin{equation}
    \textit{T}_{\mathcal{O}}^{\mathcal{J}_{EE}} =
    \textit{T}_{\mathcal{O}}^{\mathcal{J}_1}
    \textit{T}_{\mathcal{J}_1}^{\mathcal{J}_2}
    \textit{T}_{\mathcal{J}_2}^{\mathcal{J}_3}
    \textit{T}_{\mathcal{J}_3}^{\mathcal{J}_4}
    \textit{T}_{\mathcal{J}_4}^{\mathcal{J}_{EE}}
\end{equation}

\subsection{Constructing the Jacobian}
The series-parallel hybrid Jacobian is dependent on $\frac{\partial \mathbf{X}}{\partial \mathbf{Q}}$ and $\frac{\partial \mathbf{Q}}{\partial \textbf{M}}$. Here 
\textbf{X} is the first three rows in the last column of $\textit{T}_{\mathcal{O}}^{\mathcal{J}_{EE}}$. We are only concerned with the linear velocity Jacobian, as we only have 3-DoF. Differentiating the transformation matrix with respect to each motor gives the Jacobian, i.e.
\begin{equation}
\mathbf{J}(\textbf{M}) =
\begin{bmatrix}
\frac{\partial T_\mathcal{O}^{\mathcal{J}_{EE}}(1:3,4)}{\partial M_1} & \frac{\partial T_\mathcal{O}^{\mathcal{J}_{EE}}(1:3,4)}{\partial M_2} & \frac{\partial T_\mathcal{O}^{\mathcal{J}_{EE}}(1:3,4)}{\partial M_3}
\end{bmatrix} \label{eq:expexpjac}
\end{equation}
where
\begin{equation}
    \frac{\partial T_\mathcal{O}^{\mathcal{J}_{EE}}}{\partial\text{M}_i} = \sum_{j=1}^4 \frac{\partial T_\mathcal{O}^{\mathcal{J}_{EE}}}{\partial q_j} \frac{\partial q_j}{\partial\text{M}_i}   
\end{equation}
for i = 1, 2, 3. We can compute $\frac{\partial q_j}{\partial\text{M}_i}$ by differentiating the loop-closure equations. From the first loop-closure equation, denoting $\vec{OA}$ as $\textbf{V}_2$ and expanding the right-hand side gives:
\begin{equation}
\begin{split}
        l_2^2 &= [\text{P}_{2x} -\text{V}_{2x}Cq_2-\text{V}_{2z}Sq_2]^2\\ &+ [\text{V}_{2x}Sq_1Sq_2 - \text{V}_{2z}Sq_1Cq_2]^2\\ &+ [\text{P}_{2z} + \text{M}_2 - \text{V}_{2z}Cq_1Cq_2 + \text{V}_{2x}Cq_1Sq_2]^2
\end{split}
    \label{2.53}
\end{equation}
In order to take the derivative with respect to $\text{M}_1$ and $\text{M}_2$ we apply the chain rule and find the partial derivatives for the intermediate variables $q_1$, $q_2$, and $\text{M}_2$. For clarity, we will set the terms on the right-hand side equal to the following:
\begin{equation}
    \text{A}_2 = \text{P}_{2x} -\text{V}_{2x}Cq_2-\text{V}_{2z}Sq_2
    \label{2.54}
\end{equation}
\begin{equation}
    \text{B}_2 = \text{V}_{2x}Sq_1Sq_2 - \text{V}_{2z}Sq_1Cq_2
    \label{2.55}
\end{equation}
\begin{equation}
    \text{C}_2 = \text{P}_{2z} + \text{M}_2 - \text{V}_{2z}Cq_1Cq_2 + \text{V}_{2x}Cq_1Sq_2
    \label{2.56}
\end{equation}
Taking the partial derivative of \eqref{2.53} with respect to $q_1$ gives:
\begin{equation}
\begin{split}
    \text{D}_2 &= \text{B}_2(\text{V}_{2x}Cq_1Sq_2 - \text{V}_{2z}Cq_1Cq_2) \\
               &\quad + \text{C}_2(\text{V}_{2z}Sq_1Cq_2 - \text{V}_{2x}Sq_1Sq_2)
\end{split}
\end{equation}
Taking the partial derivative of \eqref{2.53} with respect to $q_2$ gives:
\begin{equation}
\begin{split}
    \text{E}_2 &= \text{A}_2(\text{V}_{2x}Sq_2 - \text{V}_{2z}Cq_2) \\
               &\quad + \text{B}_2(\text{V}_{2x}Sq_1Cq_2 + \text{V}_{2z}Sq_1Sq_2) \\
               &\quad + \text{C}_2(\text{V}_{2z}Cq_1Sq_2 + \text{V}_{2x}Cq_1Cq_2)
\end{split}
\end{equation}

The partial derivative of \eqref{2.53} with respect to $\text{M}_2$ is simply $\text{C}_2$. We can now relate $q_2$ to $\text{M}_i$ for i = 1, 2
\begin{equation}
    0 = \text{E}_2\frac{\partial q_2}{\partial\text{M}_i} + \text{D}_2\frac{\partial q_1}{\partial\text{M}_i} + \text{C}_2\frac{\partial\text{M}_2}{\partial\text{M}_i}
\end{equation}

$\frac{\partial\text{M}_2}{\partial\text{M}_1}$ = $\frac{\partial q_1}{\partial\text{M}_2}$ = 0, resulting in:
\begin{equation}
    \frac{\partial q_2}{\partial\text{M}_1} = -\frac{\text{D}_2}{\text{E}_2}\frac{\partial q_1}{\partial\text{M}_1}, \quad \frac{\partial q_2}{\partial\text{M}_2} = \frac{-\text{C}_2}{\text{E}_2}
\end{equation}

Where $\frac{\partial q_1}{\partial\text{M}_1} = 1 $ (direct-drive). Performing these operations on the other three loop-closure equations would yield all the joint-motor relationships. We can now find the partial derivatives of the homogeneous transformation matrix with respect to each motor:
\begin{equation}
\begin{split}
    \frac{\partial T_{\mathcal{O}}^{\mathcal{J}_{EE}}}{\partial\text{M}_1} &= \frac{\partial T_\mathcal{O}^{\mathcal{J}_1}}{\partial\ q_1}\frac{\partial q_1}{\partial \text{M}_1}T_{\mathcal{J}_1}^{\mathcal{J}_2}T_{\mathcal{J}_2}^{\mathcal{J}_3}T_{\mathcal{J}_3}^{\mathcal{J}_4}T_{\mathcal{J}_4}^{\mathcal{J}_{EE}} \\ &+ T_\mathcal{O}^{\mathcal{J}_1}\frac{\partial T_{\mathcal{J}_1}^{\mathcal{J}_2}}{\partial\ q_2}\frac{\partial q_2}{\partial \text{M}_1}T_{\mathcal{J}_2}^{\mathcal{J}_3}T_{\mathcal{J}_3}^{\mathcal{J}_4}T_{\mathcal{J}_4}^{\mathcal{J}_{EE}} \\ &+T_\mathcal{O}^{\mathcal{J}_1}T_{\mathcal{J}_1}^{\mathcal{J}_2}\frac{\partial T_{\mathcal{J}_2}^{\mathcal{J}_3}}{\partial\ q_3}\frac{\partial q_3}{\partial \text{M}_1}T_{\mathcal{J}_3}^{\mathcal{J}_4}T_{\mathcal{J}_4}^{\mathcal{J}_{EE}} \\ &+T_\mathcal{O}^{\mathcal{J}_1}T_{\mathcal{J}_1}^{\mathcal{J}_2}T_{\mathcal{J}_2}^{\mathcal{J}_3}\frac{\partial T_{\mathcal{J}_3}^{\mathcal{J}_4}}{\partial\ q_4}\frac{\partial q_4}{\partial \text{M}_1}T_{\mathcal{J}_4}^{\mathcal{J}_{EE}} 
    \end{split}
\end{equation}
\begin{equation}
\begin{split}
    \frac{\partial T_\mathcal{O}^{\mathcal{J}_{EE}}}{\partial\text{M}_2} &= T_\mathcal{O}^{\mathcal{J}_1}\frac{\partial T_{\mathcal{J}_1}^{\mathcal{J}_2}}{\partial\ q_2}\frac{\partial q_2}{\partial \text{M}_2}T_{\mathcal{J}_2}^{\mathcal{J}_3}T_{\mathcal{J}_3}^{\mathcal{J}_4}T_{\mathcal{J}_4}^{\mathcal{J}_{EE}} \\ &+T_\mathcal{O}^{\mathcal{J}_1}T_{\mathcal{J}_1}^{\mathcal{J}_2}\frac{\partial T_{\mathcal{J}_2}^{\mathcal{J}_3}}{\partial\ q_3}\frac{\partial q_3}{\partial \text{M}_2}T_{\mathcal{J}_3}^{\mathcal{J}_4}T_{\mathcal{J}_4}^{\mathcal{J}_{EE}} \\&+T_\mathcal{O}^{\mathcal{J}_1}T_{\mathcal{J}_1}^{\mathcal{J}_2}T_{\mathcal{J}_2}^{\mathcal{J}_3}\frac{\partial T_{\mathcal{J}_3}^{\mathcal{J}_4}}{\partial\ q_4}\frac{\partial q_4}{\partial \text{M}_2}T_{\mathcal{J}_4}^{\mathcal{J}_{EE}} 
    \end{split}
\end{equation}
\begin{equation}
\begin{split}
    \frac{\partial T_{\mathcal{O}}^{\mathcal{J}_{EE}}}{\partial\text{M}_3} &= T_\mathcal{O}^{\mathcal{J}_1}T_{\mathcal{J}_1}^{\mathcal{J}_2}\frac{\partial T_{\mathcal{J}_2}^{\mathcal{J}_3}}{\partial\ q_3}\frac{\partial q_3}{\partial \text{M}_3}T_{\mathcal{J}_3}^{\mathcal{J}_4}T_{\mathcal{J}_4}^{\mathcal{J}_{EE}} \\ &+T_\mathcal{O}^{\mathcal{J}_1}T_{\mathcal{J}_1}^{\mathcal{J}_2}T_{\mathcal{J}_2}^{\mathcal{J}_3}\frac{\partial T_{\mathcal{J}_3}^{\mathcal{J}_4}}{\partial\ q_4}\frac{\partial q_4}{\partial \text{M}_3}T_{\mathcal{J}_4}^{\mathcal{J}_{EE}} 
    \end{split}
\end{equation}
We can now find the Jacobian using Equation \ref{eq:expexpjac}.

\end{document}